\documentclass[11pt]{article}

\PassOptionsToPackage{table}{xcolor}
\usepackage{acl}

\usepackage{times}
\usepackage{latexsym}

\usepackage[T1]{fontenc}
\usepackage{amsmath}
\usepackage{pifont} 
\usepackage[utf8]{inputenc}

\usepackage{microtype}

\usepackage{inconsolata}

\usepackage{graphicx}

\usepackage{booktabs}
\usepackage{multirow}

\title{LaME: Learning to Think in Latent Space for Multimodal Embedding via Information Bottleneck}

\author{
 \textbf{Peixi Wu\textsuperscript{1}},
 \textbf{Biao Yang\textsuperscript{2}},
 \textbf{Feipeng Ma\textsuperscript{1}},
 \textbf{Bosong Chai\textsuperscript{3}},
 \textbf{Bo Lin\textsuperscript{4}},
\\
 \textbf{Wei Yuan\textsuperscript{2}},
 \textbf{Fan Yang\textsuperscript{2}},
 \textbf{Tingting Gao\textsuperscript{2}},
 \textbf{Hebei Li\textsuperscript{1}$^\dagger$},
 \textbf{Xiaoyan Sun\textsuperscript{1,5}}
\\
\\
 \textsuperscript{1}University of Science and Technology of China \quad
 \textsuperscript{2}Kuaishou Technology \quad \\
 \textsuperscript{3}Zhejiang University \quad
 \textsuperscript{4}Tsinghua University \\
\textsuperscript{5}Institute of Artificial Intelligence, Hefei Comprehensive National Science Center
\\
 \texttt{\{wupeixi,lihebei\}@mail.ustc.edu.cn}, \quad
 \texttt{sunxiaoyan@ustc.edu.cn}
\\
}

\begin{document}
\maketitle

\newcommand\blfootnote[1]{%
  \begingroup
  \renewcommand\thefootnote{}\footnote{#1}%
  \addtocounter{footnote}{-1}%
  \endgroup
}
\blfootnote{$^\dagger$Corresponding author. This work was done when Peixi Wu was an intern at Kuaishou Technology.}

\vspace{-1.5em}


\begin{abstract}
Reasoning-driven universal multimodal embedding has advanced rapidly by introducing Chain-of-Thought (CoT) reasoning into the embedding pipeline.
Despite the strong performance across both general and complex tasks, this paradigm suffers from two core limitations:
(i) autoregressive CoT reasoning incurs high computational cost,
making it impractical for low-latency retrieval;
and (ii) embedding performance is heavily coupled with CoT annotation quality, 
making large-scale training unreliable.
These raise fundamental questions: \textit{Is textual CoT the optimal form of reasoning for embedding,
and can effective embedding reasoning be accomplished in latent space?}
To this end, we propose \textbf{LaME} (\textbf{L}\textbf{a}tent Reasoning \textbf{M}ultimodal \textbf{E}mbedding),
which formulates embedding-oriented latent reasoning as a weakly supervised
information bottleneck.
LaME employs $K$ learnable reason tokens as a fixed-capacity bottleneck,
completing all reasoning within a single forward pass.
The two weak supervision signals structurally decouple contrastive from autoregressive
objectives and eliminate dependence on CoT annotations,
while a two-stage training pipeline ensures stable convergence.
Experiments on MMEB-v2 and MRMR show that LaME achieves competitive performance,
surpassing some explicit CoT-based models, while delivering
\textbf{$60{\times}$} faster inference than explicit CoT methods and
\textbf{$2{\times}$} faster than latent baselines with throughput comparable to discriminative embedding models.
The code is available at \url{https://github.com/PeppaWu/LaME}.
\end{abstract}

\section{Introduction}

\begin{figure}[t]
    \centering
    \includegraphics[width=\columnwidth]{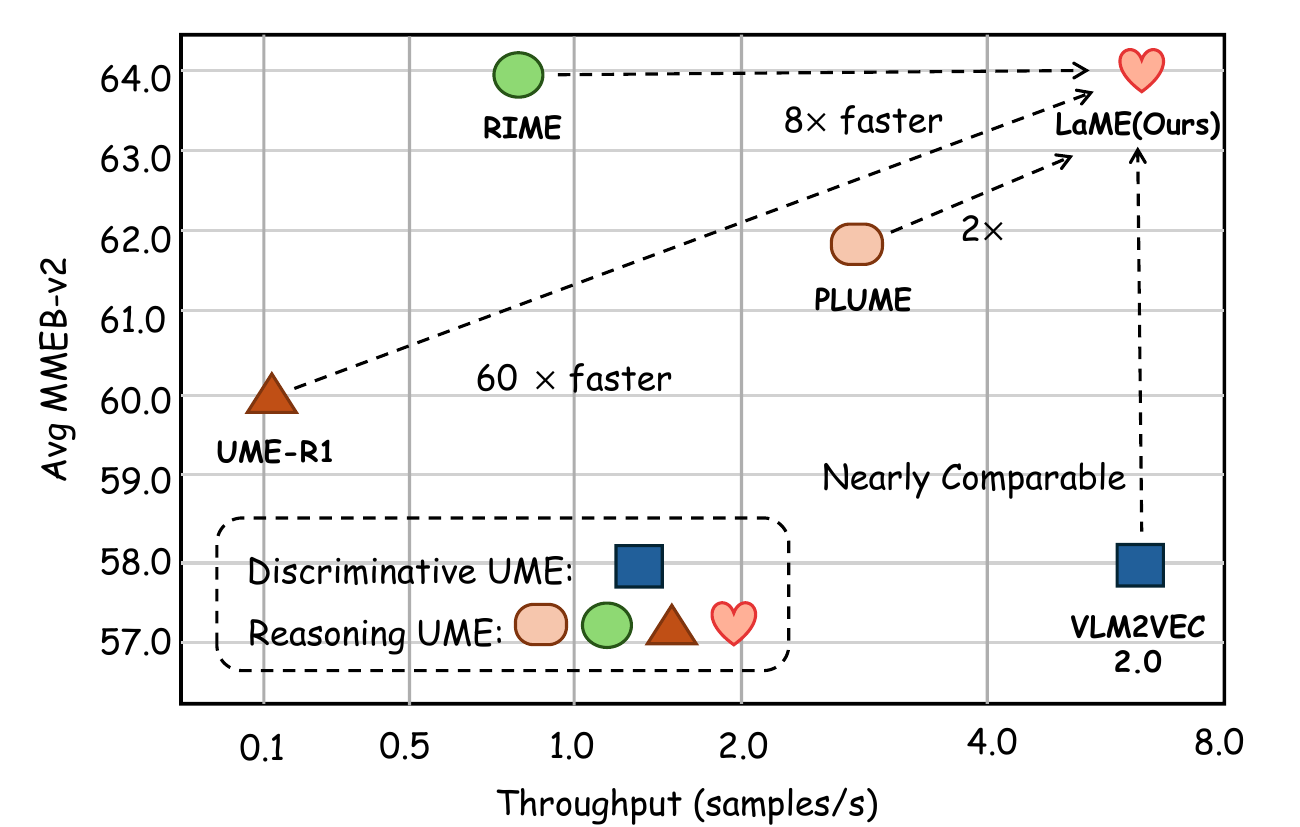}
    \caption{
    Performance vs.\ throughput on MMEB-V2 on a single GPU. LaME achieves competitive performance with reasoning-driven methods while being $60{\times}$ faster than explicit CoT approaches and $2{\times}$ faster than iterative latent approaches.
    }
    \label{fig:head}
\end{figure}


The rapid development of Multimodal Large Language Models (MLLMs)~\citep{liu2024llavanext, chen2024internvl, wang2024qwen2vl} has fundamentally transformed multimodal embedding learning.
Traditional dual encoders such as CLIP~\citep{radford2021learning} and EVA-CLIP~\citep{sun2023evaclip} align cross-modal representations via contrastive pre-training, yet struggle with complex multimodal inputs.
Recent works including VLM2Vec~\citep{jiang2024vlm2vec}, E5-V~\citep{jiang2024e5v}, GME~\citep{zhang2025gme} and so on~\citep{liu2025lamra, lin2024mmembed} demonstrate that MLLMs can be fine-tuned into powerful embedding backbones by leveraging intrinsic vision-language fusion and instruction-following capabilities, marking a paradigm shift from modality-specific encoding to unified multimodal representation learning.

Based on this, recent works further exploit the generative capacity of MLLMs
by introducing explicit Chain-of-Thought (CoT) reasoning~\citep{wei2022chain, deepseekr1}
to enrich the information available for generative embedding.
UME-R1~\citep{umer1}, RIME~\citep{wu2026rime}, RGE~\citep{liu2025rge}, and Think-Then-Embed~\citep{cui2026think} demonstrate that
generative embeddings empowered by internal or external reasoning
outperform purely discriminative ones.
However, this paradigm suffers from two core limitations
that constrain its practical feasibility.
(1) The CoT-based reasoning incurs high computational costs at the inference stage. Even with various acceleration strategies, it remains clearly impractical for large-scale and low-latency retrieval scenarios.
(2) There exists a strong coupling between embedding generation
and cold-start CoT annotation.
Defective oracle reasoning trajectories readily degrade learned embeddings,
making them highly susceptible to explicit textual annotations.
These limitations raise a deeper question:
\textit{Is textual CoT the optimal form of reasoning for embedding?}

Motivated by Coconut~\citep{hao2024coconut},
PLUME~\citep{he2026plume}  integrates latent and explicit reasoning paradigms,
iteratively feeding hidden states back as input embeddings
 to explore latent reasoning while retaining partial explicit CoT.
Its training still follows a progressive learning scheme, replacing textual CoT with latent transitions~\citep{ma2025cotvalve, deng2024implicit, zhang2025simcot, zhang2025lightthinker}. This inherently limits flexible latent reasoning, as the model is guided to imitate text-based reasoning patterns instead of learning thinking optimal for retrieval tasks.
To this end, the information bottleneck (IB)~\citep{tishby2000information, alemi2017deep, wu2025efficient} can be introduced to spark
unconstrained exploration in latent space.
By constraining \emph{what} the latent thinking must preserve through a fixed-capacity bottleneck,
IB leaves \emph{how} the model organizes its internal representations entirely unconstrained,
enabling it to spontaneously discover latent structures
without adhering to any prescribed reasoning format.

Building on this perspective, we propose \textbf{LaME} (\textbf{L}\textbf{a}tent Reasoning \textbf{M}ultimodal \textbf{E}mbedding) 
which instantiates the IB principle through $K$ learnable reason tokens
appended to the MLLM input as a fixed-capacity bottleneck.
The two weak supervision signals shape the bottleneck:
(1) a lightweight decoder head that reconstructs retrieval targets, predefined answers and keywords from the first $K_r$ latent tokens, serving as a reasoning probe that requires no CoT process at all and thus decouples training from the quality of CoT annotations;
(2) an aggregated embedding head that merges the remaining $K_e$ latent tokens
into one embedding and optimizes it against the retrieval objective,
structurally separating contrastive
from generative supervision and aligning latent reasoning with the retrieval objectives.
Together, these two heads ensure latent tokens preserve adequate information.

\begin{figure*}[!t]
    \centering
    \includegraphics[width=\textwidth]{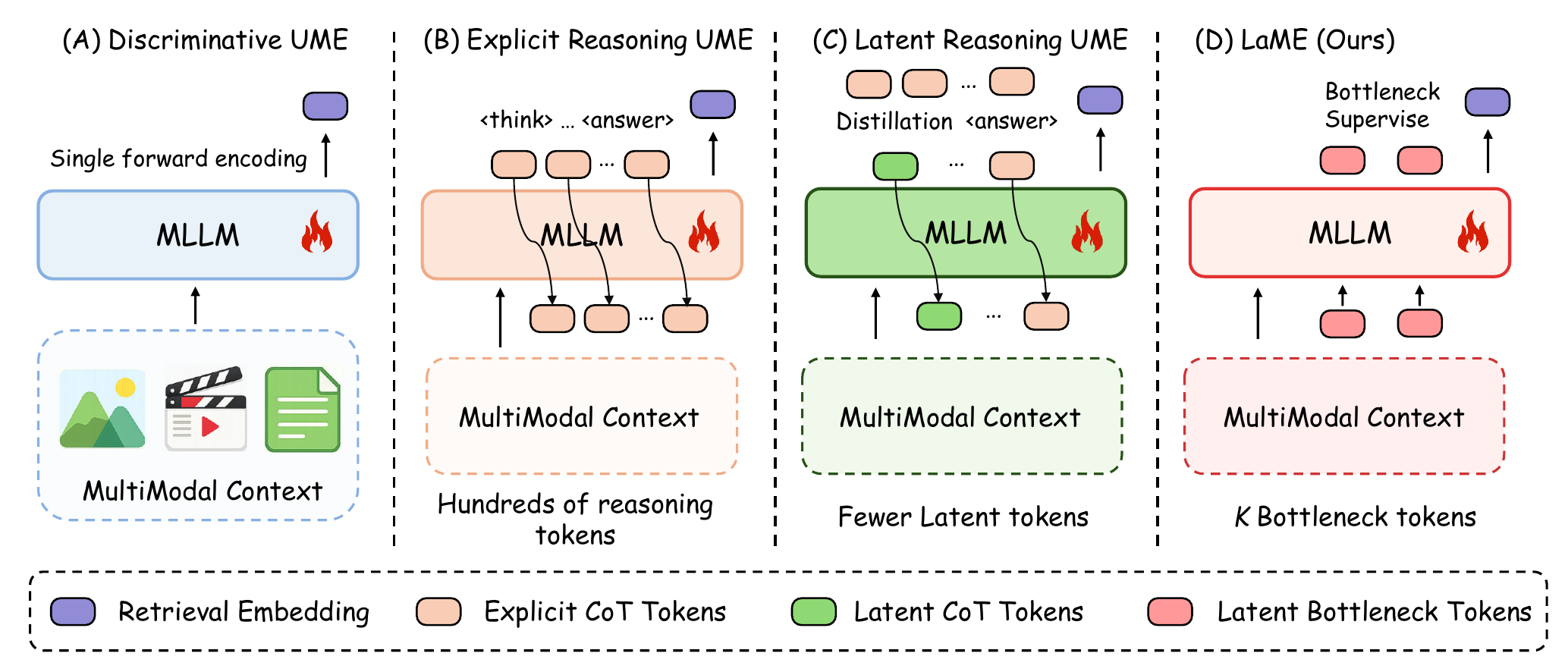}
    \caption{
    Comparison of universal multimodal embedding paradigms. 
    (A)~\textbf{Discriminative}: direct encoding without reasoning.
    (B)~\textbf{Explicit reasoning}: autoregressive CoT generation before embedding.
    (C)~\textbf{Latent reasoning}: iterative hidden-state rollout.
    (D)~\textbf{LaME}: all reasoning is confined to bottleneck tokens in a single forward pass.
    }
    \label{fig:Compare}
\end{figure*}

However, end-to-end joint training is unstable because randomly initialized reasoning tokens lack semantic structure, making IB supervision ineffective in early training.
To address this, we adopt a two-stage training pipeline. In the Bottleneck Warm-up stage, the MLLM backbone is frozen, and only the reasoning tokens and supervision heads are optimized to stabilize the bottleneck initialization. In the Joint Optimization stage, all parameters except the vision encoder are unfrozen, and an additional embedding token is appended to the end of the input for target
 contrastive supervision, enabling effective gradient propagation for embedding-oriented latent reasoning.
In summary, our contributions are summarized as follows:
\begin{itemize}
    \item We propose \textbf{LaME}, a framework that casts embedding-oriented latent reasoning as a weakly supervised information bottleneck, completely decoupling internal thinking from explicit CoT formats and exploring latent reasoning within a single forward pass.
    \item  We design dual-head bottleneck supervision that structurally separates contrastive from generative supervision, together with a two-stage training pipeline that ensures stable bottleneck convergence.
    \item By confining latent reasoning to prefill tokens within a single forward pass, LaME achieves competitive performance on MMEB-v2~\cite{meng2025vlm2vecv2} and MRMR~\cite{chen2025mrmr}, with \textbf{$60{\times}$} faster inference than explicit CoT methods and \textbf{$2{\times}$} faster inference than iterative latent rollout baselines.
\end{itemize}

\section{Related Works}

\subsection{Multimodal Embedding Models}

Early multimodal embeddings rely on dual encoders with contrastive objectives, e.g., CLIP~\citep{radford2021learning} and SigLIP~\citep{zhai2023sigmoid}.
With MLLMs, VLM2Vec~\citep{jiang2024vlm2vec}, E5-V~\citep{jiang2024e5v}, MM-Embed~\citep{lin2024mmembed}, GME~\citep{zhang2025gme}, UniME~\citep{gu2025unime}, UniME-V2~\citep{gu2026unimev2}, and RzenEmbed~\citep{jian2025rzenembed} show that MLLMs serve as strong embedding backbones via contrastive fine-tuning; MetaEmbed~\citep{metaembed2025} explores test-time scaling for multimodal retrieval.
Recently, reasoning-enhanced methods have emerged:
UME-R1~\citep{umer1} and Think-Then-Embed~\citep{cui2026think} demonstrate that explicit CoT before embedding yields substantial gains, RIME~\citep{wu2026rime} introduces retrieval-friendly rewriting, while PLUME~\citep{he2026plume} explores latent reasoning via iterative hidden-state rollouts.
However, explicit CoT methods~\citep{cui2026think, liu2026crem, wu2025spiking} entangle retrieval learning with autoregressive generation,  incurring heavy inference costs
and CoT annotation dependency.

\subsection{Latent Space Reasoning}

Chain-of-thought prompting~\citep{wei2022chain}
enables complex reasoning but incurs decoding overhead.
Coconut~\citep{hao2024coconut} and Pause tokens~\citep{goyal2024think}
pioneer latent reasoning by feeding hidden states back as inputs
or adding extra computation slots without text generation.
Subsequent works compress CoT into latent spaces:
Implicit CoT~\citep{deng2024implicit},
CODI~\citep{shen2025codi}, and CoLaR~\citep{lu2025colar}
achieve compression via variable-length sequences,
explicit-to-implicit training, self-distillation,
or dynamic single-pass schemes.
TokenSkip~\citep{xing2025tokenskip} supports step-level controllable compression.
SoftCoT~\citep{xu2025softcot}, 
SIM-CoT~\citep{zhang2025simcot}, and LightThinker~\citep{zhang2025lightthinker}
replace discrete CoT with instance-specific soft tokens.
In retrieval, LaSER~\citep{jin2026laser} and PLUME~\citep{he2026plume}
extend latent reasoning to dense retrievers and multimodal embeddings,
representing preliminary explorations for retrieval.

\begin{figure*}[!t]
    \centering
    \includegraphics[width=\textwidth]{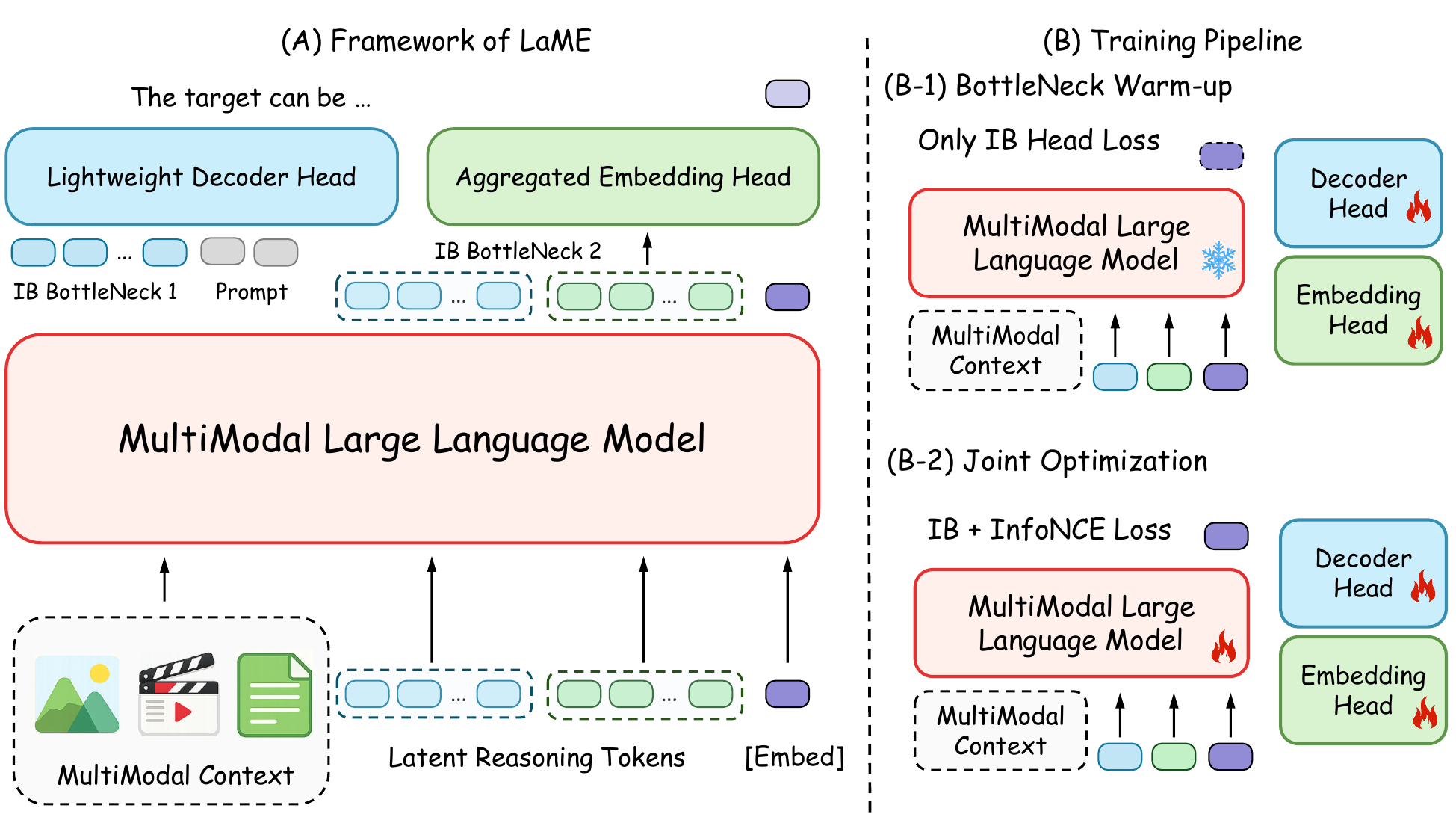}
    \caption{
    Overview of the LaME framework.
$K$ reason tokens form an information bottleneck. A Decoder Head and an Embedding Head  supervise them with two-stage  training: Bottleneck Warm-up and Joint Optimization.
    }
    \label{fig:Main}
\end{figure*}
\subsection{Information Bottleneck for Representation Learning}

The IB principle~\citep{tishby2000information} learns representations that compress input $X$ while preserving task-relevant information about $Y$, optimizing $\min I(Z,X) - \beta\, I(Z,Y)$.
Deep Variational IB~\citep{alemi2017deep} makes this tractable for neural networks; related ideas underpin CPC~\citep{oord2018representation} and Deep InfoMax~\citep{hjelm2019learning}.
Recent work shows that projection heads implicitly act as bottlenecks~\citep{zhuo2025projector}, and explicit IB objectives improve contrastive and vision-language pretraining~\citep{li2025climb,ji2025cibr,almudevar2025aligning}.
OMIB~\citep{wu2025omib} extends IB to multimodal settings with optimality guarantees.
BToks~\citep{sun2026btoks} uses learnable bottleneck tokens as pooling mechanisms with generative condensation supervision.
Our work differs in that bottleneck tokens serve as a medium for latent reasoning rather than mere pooling, supervised by an IB objective.

\section{Our Framework}

This section begins with preliminary on the background of universal multimodal embeddings (Section~\ref{sec:preliminary}).
We then introduce reason tokens as an information bottleneck
that confines latent reasoning to a single forward pass (Section~\ref{sec:ib}).
Next, we present dual-head supervision
that steers the bottleneck without relying on CoT  (Section~\ref{sec:dualhead}).
Finally, we introduce the two-stage training pipeline
for stable bottleneck convergence (Section~\ref{sec:twostage}).

\subsection{Preliminary}
\label{sec:preliminary}

Given a query $q$ and a candidate set,
the goal of Universal Embedding Model is to retrieve
the most relevant target $t^+$
via similarity in a shared embedding space.
Existing methods differ in the thinking process
injected before embedding extraction.
\textbf{Discriminative} methods directly encode the input
into a normalized embedding $f_\theta(x)$
without introducing any intermediate thinking process,
and optimize the resulting representations with InfoNCE:
\begin{equation}
\mathcal{L}_{\text{NCE}} = -\frac{1}{N} \sum_{i=1}^{N}
\log \frac{\exp(f_\theta(q_i) \cdot f_\theta(t_i^+) / \tau)}
{\sum_{j=1}^{N} \exp(f_\theta(q_i) \cdot f_\theta(t_j) / \tau)},
\end{equation}
where $\tau$ is the temperature.
It is efficient but lacks intermediate reasoning for complex queries.
\textbf{Explicit reasoning} methods address this
by first generating a textual chain-of-thought $o^x = g_\phi(x)$,
then embedding the concatenation $f_\theta(x, o^x)$.
While effective, the autoregressive decoding of $o^x$
is prohibitively slow at inference time.
\textbf{Latent reasoning} methods replace partial textual thinking
with hidden state iteration,
but retain inference overhead
and rely on retrieval-friendly CoT annotations
for progressive imitation training.
Our method takes a different route:
it confines all latent reasoning to $K$ learnable tokens
within a single forward pass,
governed by the IB objective.

\subsection{Reason Tokens as Information Bottleneck}
\label{sec:ib}
The core idea of LaME is to append $K$ learnable reason tokens
to the input and supervise them
with an information bottleneck (IB) objective,
thereby guiding the reasoning process indirectly.
Concretely, given an input $\mathbf{x}$
which may contain interleaved text and visual tokens,
we append $K$ reason tokens
$\mathbf{r} = [r_1, \ldots, r_K]$
to form the extended input $[\mathbf{x}; \mathbf{r}]$.
A single forward pass through the MLLM
produces hidden states for all positions:
\begin{equation}
[\mathbf{h}_\mathbf{x}; \mathbf{h}_\mathbf{r}] = \text{MLLM}([\mathbf{x}; \mathbf{r}]).
\end{equation}
Under causal attention,
the reason tokens $\mathbf{h}_\mathbf{r}$ attend to all preceding context in $\mathbf{x}$
and aggregate task-relevant information.
However, the final embedding is derived from the full hidden states,
allowing it to spontaneously attend to
either the reason tokens or the original input.

The bottleneck lies in the supervision of $\mathbf{h}_\mathbf{r}$:
we constrain the reason tokens via dedicated heads.
Under this constraint, the reason tokens must think and abstract task-relevant cues from the input,
while the finite capacity of $K$ tokens forces them
to retain only critical information and discard noise.

From an information-theoretic perspective,
this instantiates the IB principle~\citep{tishby2000information}.
Let $Z$ denote the representation carried by $\mathbf{h}_\mathbf{r}$,
$X$ the input, and $Y$ the target defined by the supervision heads.
The supervision heads minimize $I(Z; X)$
via the fixed capacity of $K$ tokens
while maximizing $I(Z; Y)$ via retrieval-oriented losses.
Unlike variational IB, LaME requires no explicit rate regularization;
the finite token count itself is the hard capacity constraint.

\subsection{Dual-Head Supervision}
\label{sec:dualhead}

We supervise the reason tokens with two weak IB supervision heads
that address the two core limitations of explicit reasoning methods:
dependency on explicit CoT annotation and misalignment between
the reasoning and the retrieval objective.


\paragraph{Decoder Head (Reasoning Probe).}
We attach a lightweight autoregressive decoder
that decodes target-relevant content from $\mathbf{h}_r$.
Specifically, the hidden states of the first $K_r$ reason tokens
are used to reconstruct retrieval targets,
predefined answers or keywords.
In practice, we reuse the answer fields
from existing research datasets~\cite{jiang2026embedrl} as supervision targets:
\begin{equation}
\mathcal{L}_{\text{Dec}} = -\sum_{t=1}^{|y|}
\log p_\psi(y_t \mid y_{<t}, \mathbf{h}_r^{1:K_r}, s),
\end{equation}
where $\psi$ denotes the decoder parameters,
$s$ is a decode prompt,
and $y$ is the reconstruction target.
This head acts as a reasoning probe:
if the latent thinking fails to distill useful information,
decoding quality degrades,
automatically providing gradient signals to guide reasoning optimization.
Crucially, the supervision uses only the target as a weak signal
without supervising intermediate thinking steps, 
decoupling training from CoT annotation.

\paragraph{Embedding Head (Retrieval Alignment).}
The remaining $K_e$ reason token hidden states
where $K_r + K_e = K$
are aggregated via mean pooling and normalized
to produce the reason embedding:
\begin{equation}
\mathbf{e} = \text{Norm}\left(
\frac{1}{K_e}\sum_{k=K_r+1}^{K} \mathbf{h}_r^{k}
\right).
\end{equation}
This embedding is trained with InfoNCE:
\begin{equation}
\mathcal{L}_{\text{Emb}} = -\frac{1}{N}\sum_{i=1}^{N}
\log \frac{\exp(\mathbf{e}_{q_i} \cdot \mathbf{e}_{t_i^+} / \tau)}
{\sum_{j=1}^{N}\exp(\mathbf{e}_{q_i} \cdot \mathbf{e}_{t_j} / \tau)}.
\end{equation}
Since this head bypasses the autoregressive generation path entirely,
it structurally separates contrastive from generative optimization,
ensuring that the latent reasoning process
remains oriented toward the retrieval objective.

\paragraph{Joint Objective.}
In addition to the two IB heads,
we append a learnable embed token after the reason tokens.
Its hidden state is projected and normalized
as the final retrieval embedding,
trained with InfoNCE.
The overall loss is:
\begin{equation}
\mathcal{L} = \mathcal{L}_{\text{NCE}}
+ \lambda_{\text{Dec}} \mathcal{L}_{\text{Dec}}
+ \lambda_{\text{Emb}} \mathcal{L}_{\text{Emb}}
+ \lambda_{\text{Div}} \mathcal{L}_{\text{Div}},
\end{equation}
where $\lambda_{\text{Dec}}$, $\lambda_{\text{Emb}}$, $\lambda_{\text{Div}}$
are balancing hyperparameters,
and $\mathcal{L}_{\text{Div}}$ is a diversity regularizer
penalizing pairwise cosine similarity among reason tokens:
\begin{equation}
\mathcal{L}_{\text{Div}} =
\frac{1}{K(K{-}1)}
\sum_{i \neq j}
\left| \cos\!\left(\mathbf{h}_r^i, \mathbf{h}_r^j\right) \right|,
\end{equation}
which penalizes representational redundancy
and encourages each reason token to attend to
distinct facets of the input during latent reasoning.

\subsection{Two-Stage Training}
\label{sec:twostage}

Directly optimizing $\mathcal{L}$ end-to-end is unstable.
Randomly initialized $\mathbf{r}$ carries no semantic information,
making the IB supervision prone to failure:
$\mathcal{L}_{\text{Dec}}$ struggles to decode meaningful content
from uninformative $\mathbf{h}_r$,
and $\mathcal{L}_{\text{Emb}}$ lacks a reliable anchor
for contrastive alignment,
hindering bottleneck convergence.
We therefore adopt a two-stage training pipeline to progressively aligns these latents as shown in Figure~\ref{fig:Main}.

\paragraph{Stage 1: Bottleneck Warm-up.}
The MLLM backbone is frozen;
only the reason tokens $\mathbf{r}$ and both IB heads are optimized.
$\mathcal{L}_{\text{Dec}}$ establishes a mapping
from $\mathbf{h}_r$ to target semantics,
$\mathcal{L}_{\text{Emb}}$ provides initial retrieval-oriented structure,
and $\mathcal{L}_{\text{Div}}$ prevents early collapse.
After this stage, $\mathbf{h}_r$ already encodes meaningful information,
ready to supply stable gradients to the backbone.

\paragraph{Stage 2: Joint Optimization.}
All parameters except the vision encoder are unfrozen for optimization. The contrastive loss $\mathcal{L}_{\text{NCE}}$ associated with the embed token is introduced.
The two IB heads propagate high-quality gradients to the backbone network through the reasoning hidden states $\mathbf{h}_r$.
Meanwhile, $\mathcal{L}_{\text{NCE}}$ directly optimizes the final retrieval embeddings within the output space of the backbone, facilitating collaborative adaptation of the entire model.

This progressive scheme ensures that
by the time the backbone begins updating,
the bottleneck already carries informative signals,
preventing degenerate solutions and accelerating convergence.

\begin{table*}[!t]
\caption{Results on the MMEB-V1 benchmark~\cite{jiang2024vlm2vec}. The scores are averaged per meta-task. \textbf{Bold} denotes the best performance within each category.  Overall denotes the average score across 36 subtasks.}
\centering
\setlength{\tabcolsep}{10pt}   
\renewcommand{\arraystretch}{1.0}
\resizebox{\textwidth}{!}{
\begin{tabular}{lc cccc ccc}
\toprule
\multirow{2}{*}{\textbf{Models}} 
& \multirow{2}{*}{\textbf{Backbone}}
& \multicolumn{4}{c}{\textbf{Per Meta-Task Score}}
& \multicolumn{3}{c}{\textbf{Average Score}} \\
\cmidrule(lr){3-6} \cmidrule(lr){7-9}
& & \textbf{Classification} & \textbf{VQA} & \textbf{Retrieval} & \textbf{Grounding} & \textbf{IND} & \textbf{OOD} & \textbf{Overall} \\
\midrule
\textbf{\# of Datasets $\rightarrow$} &
& 10 & 10 & 12 & 4 & 20 & 16 & 36 \\
\midrule
\rowcolor[HTML]{EDEDED}
\multicolumn{9}{c}{\emph{Discriminative UME}} \\
CLIP & CLIP-L-0.4B &
42.8 & 9.1 & 53.0 & 51.8 & 37.1 & 38.7 & 37.8 \\
GME & Qwen2VL-2B &
54.4 & 29.9 & 66.9 & 55.5 & 49.2 & 55.2 & 51.9 \\
VLM2Vec &  Qwen2VL-2B &
58.7 & 49.3 & 65.0 & 72.9 & 64.8 & 53.4 & 59.7 \\
VLM2Vec-V2 & Qwen2VL-2B &
62.9 & 56.3 & 69.5 & 77.3 & 68.8 & 60.1 & 64.9 \\
Ops-MM-Embed & Qwen2VL-2B &
\textbf{68.1} & 65.1 & 69.2 & 80.9 & 71.8 & \textbf{65.6} & 69.0 \\
ReMatch & Qwen2VL-2B &
65.8 & \textbf{65.9} & \textbf{70.1} & \textbf{83.3} & \textbf{72.8} & 64.7 & \textbf{69.2} \\
\midrule
\rowcolor[HTML]{EDEDED}
\multicolumn{9}{c}{\emph{Explicit Reasoning UME}} \\
UME-R1 & Qwen2VL-2B &
64.8 & 62.8 & 67.6 & 77.2 & 71.5 & 60.4 & 66.6 \\
RIME & Qwen2VL-2B &
67.9 & 64.4 & 69.8 & 82.1  & 64.5 & 72.9 & 69.2 \\
Embed-RL  & Qwen2.5VL-3B &
62.8 &
67.9 &
68.6 &
90.4 & 	65.9 & 71.9 & 69.2 \\
RGE & Qwen3VL-2B &
 64.4 & 67.8 & 70.2 & 90.1 & 75.3 & 63.7 & 70.1 \\
Think-When-Needed & Qwen3VL-4B &
 68.6 & 71.7 & 68.0 & 87.0 & 75.4 & 66.2	& 71.3 \\
Think-Then-Embed & Qwen2VL-2B &
\textbf{69.7} & \textbf{72.4} & \textbf{74.0} & \textbf{90.6} & \textbf{80.5} & \textbf{67.0} & \textbf{74.5} \\
\midrule
\rowcolor[HTML]{EDEDED}
\multicolumn{9}{c}{\emph{Latent Reasoning UME}} \\

PLUME & Qwen2VL-2B &
66.1 &	59.5 &	67.6 &	79.9 &	69.6 &	62.2 &	66.3
 \\
\rowcolor[HTML]{FFF0EB}
\textbf{LaME (Ours)} & Qwen2VL-2B &
\textbf{67.6} & \textbf{66.2} & \textbf{70.5} & \textbf{81.2} & \textbf{73.1} & \textbf{64.4} & \textbf{69.3} \\
\bottomrule
\end{tabular}
}
\label{tab:main_image}
\end{table*}

\begin{table*}[!t]
\caption{Results on the full MMEB-V2 benchmark~\cite{meng2025vlm2vecv2}. CLS: classification, QA: question answering, RET: retrieval, GD: grounding, MRET: moment retrieval, VDR: ViDoRe, VR: VisRAG, OOD: out-of-distribution. Best performance per model size in \textbf{bold}. All denotes the average score across 78 subtasks.}
\centering
\renewcommand{\arraystretch}{1.06}
\resizebox{\textwidth}{!}{
\begin{tabular}{l ccccc ccccc ccccc c}
\toprule
\multirow{2}{*}{\textbf{Model}} 
& \multicolumn{5}{c}{\textbf{Image}} 
& \multicolumn{5}{c}{\textbf{Video}} 
& \multicolumn{5}{c}{\textbf{VisDoc}}
& \multirow{2}{*}{\textbf{All}}\\
\cmidrule(lr){2-6} \cmidrule(lr){7-11} \cmidrule(lr){12-16}
& \textbf{CLS} & \textbf{QA} & \textbf{RET} & \textbf{GD} & \textbf{Avg.} 
& \textbf{CLS} & \textbf{QA} & \textbf{RET} & \textbf{MRET} & \textbf{Avg.} 
& \textbf{VDRv1} & \textbf{VDRv2} & \textbf{VR} & \textbf{OOD} & \textbf{Avg.} & \\
\midrule
\textbf{\# of Datasets}
& 10 & 10 & 12 & 4 & 36 
& 5 & 5 & 5 & 3 & 18 
& 10 & 4 & 6 & 4 & 24
& 78 
\\
\midrule
\rowcolor[HTML]{EDEDED}
\multicolumn{17}{c}{\emph{$\sim$ 2B Model Size}} \\
VLM2Vec &
58.7 & 49.3 & 65.0 & 72.9 & 59.7 & 
33.4 & 30.5 & 20.6 & 33.0 & 29.0 & 
49.8 & 13.5 & 51.8 & 33.5 & 41.6 & 
47.0 \\
VLM2Vec-V2 &
62.0 & 56.3 & 69.5 & 77.3 & 64.9 & 
39.3 & 34.3 & 28.8 & 38.5 & 34.9 & 
75.5 & 44.9 & 79.4 & 39.4 & 65.4 & 
58.0 \\
GME &
54.4 & 29.9 & 66.9 & 55.5 & 51.9 & 
34.9 & 42.0 & 25.6 & 32.4 & 33.9 & 
\textbf{86.1} & \textbf{54.0} & 82.5 & 43.1 & \textbf{72.7} & 
54.1 \\
LamRA &
59.2 & 26.5 & 70.0 & 62.7 & 54.1 & 
39.3 & 42.6 & 24.3 & 34.6 & 35.2 & 
22.0 & 11.5 & 37.4 & 21.0 & 23.9 & 
40.4 \\
UME-R1 &
64.8 & 62.8 & 67.6 & 77.2 & 66.6 &
44.3 & 51.0 & 32.9 & 39.7 & 42.2 & 
72.4 & 46.2 & 79.2 & 28.9 & 62.5 & 
59.7 \\
Ops-MM-Embed &
\textbf{68.1} & 65.1 & 69.2 & 80.9 & 69.0 & 
\textbf{53.6} & \textbf{55.7} & \textbf{41.8} & 33.7 & \textbf{47.6} & 
76.4 & 53.2 & 77.6 & 32.6 & 65.5 & 
63.0 \\
PLUME &
66.5 & 59.2 & 67.6 & 79.7 & 66.3 & 
45.0 & 52.3 & 33.5 & \textbf{46.7} & 44.1 & 
72.1 & 49.8 & 78.1 & \textbf{57.4} & 67.5 & 
61.6 \\
\rowcolor[HTML]{FFF0EB}
\textbf{LaME (Ours)} &
67.6 & \textbf{66.2} & \textbf{70.5} & \textbf{81.2} & \textbf{69.3} &
45.6 & 52.3 & 36.3 & 43.5 & 44.5 &
78.2 & 48.4 & \textbf{82.8} & 55.4 & 72.1 & 
\textbf{64.4} \\

\midrule
\rowcolor[HTML]{EDEDED}
\multicolumn{17}{c}{\emph{$\sim$ 7B Model Size}} \\
ColPali-V1.3 & 
40.5 & 11.5 & 48.2 & 39.8 & 34.9 & 
26.6 & 37.8 & 21.5 & 26.2 & 28.2 & 
84.6 & 54.8 & 81.0 & 33.1 & 70.1 & 
44.2 \\

GME & 
57.7 & 34.7 & 71.2 & 59.3 & 56.0 & 
37.4 & 50.4 & 28.4 & 38.2 & 38.6 & 
\textbf{89.4} & 55.6 & 85.0 & 44.4 & 75.2 & 
57.8 \\
LamRA & 
51.7 & 34.1 & 66.9 & 56.7 & 52.4 & 
32.9 & 42.6 & 23.2 & 37.6 & 33.7 & 
56.3 & 33.3 & 58.2 & 40.1 & 50.2 & 
47.4 \\
VLM2Vec & 
62.7 & 56.9 & 69.4 & 82.2 & 65.5 & 
39.1 & 30.0 & 29.0 & 40.6 & 34.0 & 
56.9 & 9.4 & 59.1 & 38.1 & 46.4 & 
52.3 \\
CAFe & 
63.6 & 61.7 & 69.1 & \textbf{87.6} & 67.6 & 
35.8 & 58.7 & 34.4 & 39.5 & 42.4 & 
70.7 & 49.6 & 79.5 & 38.1 & 63.9 & 60.6 \\
UME-R1 & 
67.1 & 69.2 & 71.9 & 84.9 & 71.3 &
48.6 & 60.7 & 38.2 & 39.3 & 47.5 & 
75.7 & 50.5 & 83.7 & 37.6 & 67.1 & 
64.5 \\
Ops-MM-Embed & 
69.7 & 69.6 & 73.1 & 87.2 & 72.7 & 
\textbf{59.7} & 62.2 & \textbf{45.7} & 43.2 & \textbf{53.8} & 
80.1 & \textbf{59.6} & 79.3 & 67.8 & 74.4 & 
\textbf{68.9} \\
\rowcolor[HTML]{FFF0EB}
\textbf{LaME (Ours)} & 
\textbf{70.0} & \textbf{71.0} & \textbf{73.1} & 85.8 & \textbf{73.0} &
51.0 & \textbf{64.8} & 40.3 & \textbf{44.6} & 50.8 &
81.5 & 58.2 & \textbf{85.0} & \textbf{68.2} & \textbf{75.9} & 
68.8 \\

\bottomrule
\end{tabular}
}
\label{tab:main_full}
\end{table*}

\begin{table*}[!t]
\caption{Results on the MRMR benchmark~\cite{chen2025mrmr} including Art, Medicine (Med.), Science (Sci.), Humanities (Hum.), Math, Physics (Phy.), Engineering (Eng.), Business (Bus.), Negation (Neg.), Design (Des.), and Traffic (Tra.). All denotes the average score across 11 subtasks.}
\label{tab:mrmr}
\centering
\scriptsize
\setlength{\tabcolsep}{4.5pt}
\renewcommand{\arraystretch}{1.05}

\resizebox{\textwidth}{!}{%
\begin{tabular}{llc cccc cccc ccc c}
\toprule
\multirow{2}{*}{\textbf{Model}} 
& \multirow{2}{*}{\textbf{Backbone}}
& \multirow{2}{*}{\textbf{Size}}
& \multicolumn{4}{c}{\textbf{Knowledge}} 
& \multicolumn{4}{c}{\textbf{Theorem}} 
& \multicolumn{3}{c}{\textbf{Contradiction}} 
& \multirow{2}{*}{\textbf{All}} \\
\cmidrule(lr){4-7} \cmidrule(lr){8-11} \cmidrule(lr){12-14}
& & & Art & Med. & Sci. & Hum. & Math & Phy. & Eng. & Bus. & Neg. & Des. & Tra. & \\
\midrule
\rowcolor[HTML]{EDEDED}
\multicolumn{15}{c}{\emph{Explicit Reasoning UME}} \\
UME-R1 & Qwen2-VL & 7B &
77.8 & 55.7 & 72.9 & 64.1 & 27.2 & 39.2 & 32.2 & 47.8 & 7.5 & 61.9 & 41.7 & 48.0 \\
RIME & Qwen2-VL & 7B &
76.8 & 58.5 & 73.6 & 71.4 & 29.2 & 43.0 & 35.6 & 51.8 & 8.5 & 64.1 & 39.5 & 50.2 \\
\midrule
\rowcolor[HTML]{EDEDED}
\multicolumn{15}{c}{\emph{Discriminative UME \& Latent Reasoning UME}} \\
EVA-CLIP & EVA-ViT & 0.4B &
10.2 & 13.5 & 26.1 & 12.9 & 6.2 & 10.5 & 9.3 & 11.7 & 8.5 & 4.4 & 5.4 & 10.8 \\
OpenCLIP & ViT-G/14 & 1B &
56.0 & 17.9 & 33.2 & 22.0 & 5.7 & 5.0 & 7.0 & 9.7 & 13.0 & 8.1 & 12.4 & 17.3 \\
VISTA & Qwen2-VL & 2B &
21.3 & 27.8 & 32.6 & 17.0 & 18.8 & 17.1 & 17.3 & 28.6 & 20.0 & 20.2 & 9.4 & 20.9 \\
E5-V & LLaVA-Next & 8B &
25.1 & 11.7 & 16.6 & 10.8 & 2.1 & 3.4 & 2.5 & 5.2 & 11.5 & 3.7 & 2.1 & 8.6 \\
VLM2Vec & Qwen2-VL & 7B &
53.5 & 22.4 & 36.7 & 24.0 & 2.1 & 2.8 & 2.8 & 2.9 & 11.5 & 5.6 & 18.3 & 18.1 \\
ColPali & PaliGemma & 3B & 36.1 & 29.9 & 42.7 & 29.2 & 5.7 & 14.8 & 12.0 & 24.6 & \textbf{28.5} & 19.4 & 18.2 & 23.7 \\
GME & Qwen2-VL & 7B &
54.3 & 40.1 & 46.8 & 45.6 & 28.8 & 36.0 & 30.2 & 45.1 & 15.0 & 26.3 & 29.6 & 36.2 \\
MM-Embed & NV-Embed & 8B &
65.6 & 53.0 & 63.5 & 62.8 & 23.6 & 30.8 & 27.4 & 44.9 & 7.0 & 23.8 & 34.9 & 39.8 \\
Ops-MM-Embed & Qwen2-VL & 7B &
\textbf{79.3} & 52.5 & 70.0 & \textbf{67.8} & 27.7 & 39.5 & 30.1 & \textbf{52.3} & 8.0 & 55.9 & \textbf{45.8} & 48.1 \\
\rowcolor[HTML]{FFF0EB}
\textbf{LaME (Ours)} & Qwen2-VL & 7B &
73.4 & \textbf{58.2} & \textbf{73.8} & 65.6 & \textbf{29.5} & \textbf{44.4} & \textbf{36.4} & 52.2 & 8.5 & \textbf{64.9} & 40.9 & \textbf{49.8} \\
\bottomrule
\end{tabular}%
}
\end{table*}
\section{Experiments}


\subsection{Implementation Details}

Following VLM2Vec-V2, we build LaME on Qwen2-VL-2B/7B-Instruct,
with Qwen3-0.6B as the default decoder head and mean-pooling as the embedding head.
We train on 1.5 million samples
comparable to baselines,
where the decoder head reuses answer fields 
without any CoT annotation.
We only use $K{=}8$ reason tokens, with $K_r{=}4$ for decoder head and $K_e{=}4$ for embedding head.
In Stage~1, the backbone is frozen
and only the reason tokens and IB heads are warmed up for about 2000 steps.
In Stage~2, all parameters are unfrozen
and jointly trained for about 3200 steps.
The loss weights are
$\lambda_{\text{Dec}}{=}1.0$, $\lambda_{\text{Emb}}{=}1.0$,
$\lambda_{\text{Div}}{=}0.05$, and the temperature $\tau{=}0.02$.
The batch size is 512 achieved through gradient accumulation. We use the AdamW optimizer with a learning rate of $1\mathrm{e}{-5}$.


\subsection{Datasets and Metrics}

To evaluate the retrieval capability of LaME,
we adopt MMEB-V2,
a recently proposed benchmark for multimodal universal retrieval.
As shown in Table~\ref{tab:main_full},
MMEB-V2 contains 36 and 18 sub-datasets
for image and video scenarios respectively,
covering classification, VQA, retrieval, visual grounding,
moment retrieval, and so on 
with Hit@1 as the evaluation metric.
For the Visual Document scenario 
which involves document-as-image tasks,
we evaluate on 24 datasets
from ViDoRe and VisRAG,
with NDCG@5 as the evaluation metric.

To further validate the generalization of LaME
in reasoning-intensive scenarios,
we adopt MRMR,
a benchmark specifically designed to assess retrieval capability
in expert-level, reasoning-intensive tasks.
Following their standard settings,
we use nDCG@10 as the main metric.

\subsection{Main Results}
\paragraph{Results on MMEB-V1.}
As shown in Table~\ref{tab:main_image},
LaME achieves the best overall score of 69.3 among latent reasoning methods,
surpassing PLUME by 3.0 points.
It also outperforms competitive discriminative UME baselines
such as ReMatch, demonstrating that latent reasoning can advance beyond discriminative models
without incurring extra reasoning cost. 
Compared to explicit reasoning methods, LaME matches RIME and EmbedRL at 69.2, while remaining 5.2 points below Think-Then-Embed, which leverages an external 72B reasoner to generate explicit CoT.
LaME improves over PLUME on all four meta-tasks,
indicating that the information bottleneck yields consistent gains across diverse task types.
\begin{table}[!t]
\caption{Ablation on core components on MMEB-V2. All crosses
(\ding{55}\ding{55}\ding{55}) denote discriminative baseline.}
\centering
\renewcommand{\arraystretch}{1.15}
\resizebox{\columnwidth}{!}{
\begin{tabular}{ccc ccc c}
\toprule
\textbf{Dec. Head} & \textbf{Emb. Head} & \textbf{Two-Stage}
& \textbf{Image} & \textbf{Video} & \textbf{VisDoc} & \textbf{All} \\
\midrule
\ding{55} & \ding{55} & \ding{55} & 68.5 & 43.9 & 71.3 & 63.8 \\
\ding{51} & \ding{51} & \ding{55} & 68.0 & 42.9 & 70.9 & 63.3 \\
\ding{51} & \ding{55} & \ding{51} & 68.8 & 44.2 & 71.2 & 64.0 \\
\ding{55} & \ding{51} & \ding{51} & 69.0 & 44.0 & 71.7 & 64.1 \\
\rowcolor[HTML]{FFF0EB}
\ding{51} & \ding{51} & \ding{51} & \textbf{69.3} & \textbf{44.5} & \textbf{72.1} & \textbf{64.4} \\
\bottomrule
\end{tabular}
}
\label{tab:abla_com}
\end{table}

\begin{table}[!t]
\caption{Effect of the number of reason tokens $K$ on MMEB-V2 and MRMR. Throughput is measured on a single GPU.}
\centering
\renewcommand{\arraystretch}{1.05}
\resizebox{\columnwidth}{!}{
\begin{tabular}{c ccc c c}
\toprule
$K$ & \textbf{Image} & \textbf{Video} & \textbf{VisDoc} & \textbf{MRMR} & \textbf{Throughput(samples/s)} \\
\midrule
0  & 68.5 & 43.9 & 71.3 & 48.0 & $6.5 \pm 0.1$ \\
2  & 68.3 & 44.0 & 71.5 & 48.1 & $6.4 \pm 0.2$ \\
4  & 68.9 & 44.1 & 71.8 & 49.5 & $6.3 \pm 0.3$ \\
\rowcolor[HTML]{FFF0EB}
8  & \textbf{69.3} & \textbf{44.5} & \textbf{72.1} & 49.8 & $6.2 \pm 0.3$ \\
16 & 68.7 & 44.1 & 71.7 & \textbf{49.9} & $6.1 \pm 0.4$ \\
\bottomrule
\end{tabular}
}
\label{tab:abla_k}
\end{table}

\paragraph{Results on MMEB-V2.}
Table~\ref{tab:main_full} extends the evaluation to the full MMEB-V2 benchmark
covering image, video, and visual document modalities.
Among 2B models, LaME ranks first overall at 64.4,
leading Ops-MM-Embed by 1.4 and PLUME by 2.8 points.
At the 7B scale, LaME scores 68.8, trailing Ops-MM-Embed by only 0.1 points,
while achieving the best image and visual document averages.
It confirms that the latent reasoning can generalize across modalities and model scales.

\paragraph{Results on MRMR.}
Table~\ref{tab:mrmr} presents results on the MRMR benchmark.
At the 7B scale, LaME achieves an overall score of 49.8.
It ranks first on four theorem-reasoning subtasks:
Science 73.8, Math 29.5, Physics 44.4, and Engineering 36.4.
Its overall score is competitive with RIME at 50.2 and surpasses Ops-MM-Embed at 48.1.
This shows that latent reasoning remains effective on expert-level and reasoning-intensive tasks, further validating its generality for complex queries.
\begin{figure}[!t]
    \centering
    \includegraphics[width=\columnwidth]{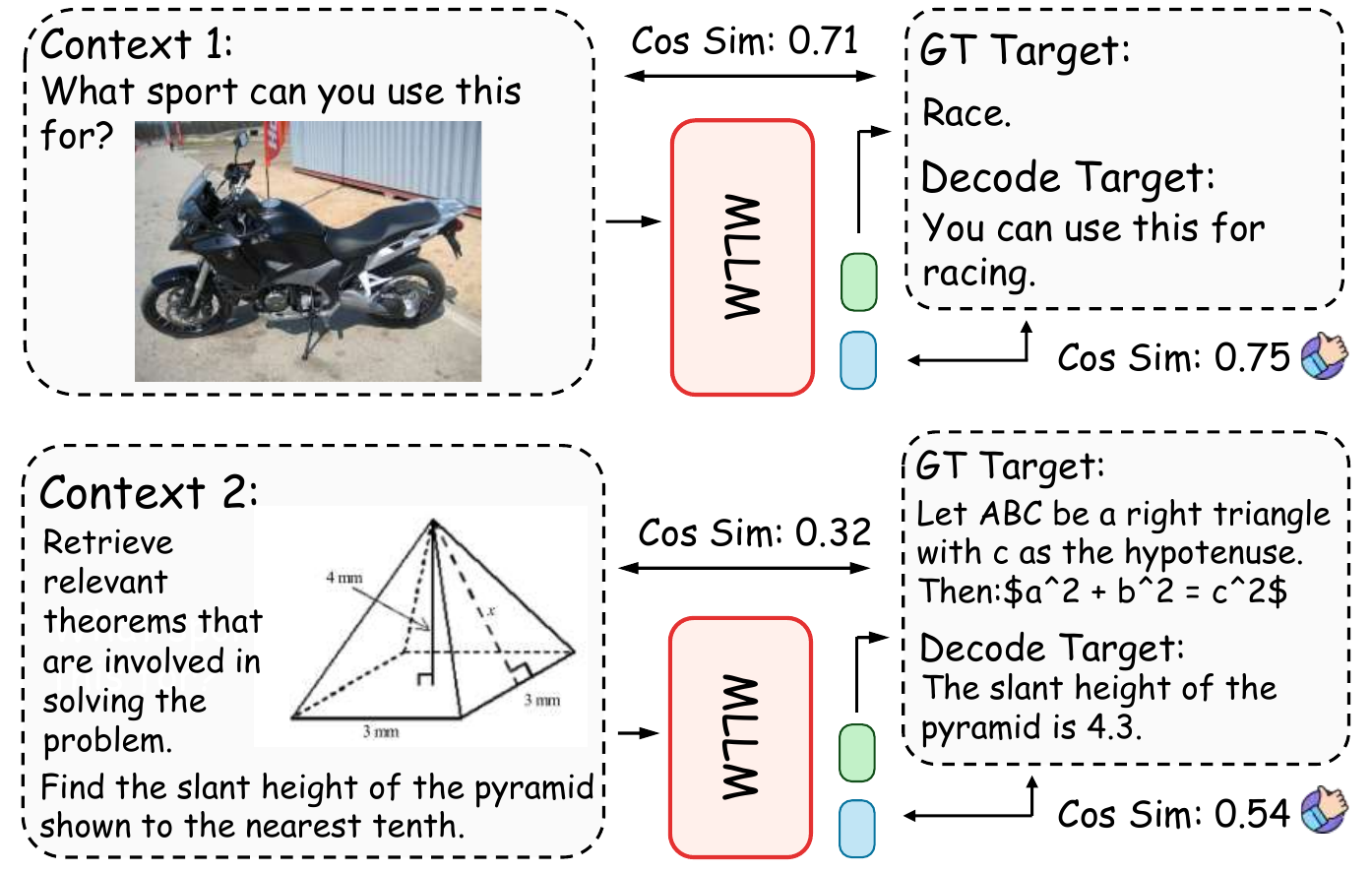}
    \caption{
    Qualitative examples of LaME. The green and blue block denotes the latent reason tokens for decoding and embedding. 
    }
    \label{fig:demo}
\end{figure}

\subsection{Ablation Studies}
\paragraph{Ablation on Core Components.}
As shown in Table~\ref{tab:abla_com},
the discriminative baseline scores 63.8 overall.
Adding both IB heads without two-stage training degrades performance to 63.3,
showing that naive attachment without careful optimization can be harmful.
With two-stage training, either the decoder head or the embedding head alone surpasses the baseline,
yielding 64.0 and 64.1 respectively.
The full configuration achieves the best result of 64.4 across all modalities,
demonstrating that all  components are jointly essential.

\paragraph{Ablation on Latent steps $K$.}
As shown in Table~\ref{tab:abla_k}, increasing $K$ from 0 to 8 steadily boosts performance across all MMEB-V2 modalities, with Image scores rising from 68.5 to 69.3, Video from 43.9 to 44.5, VisDoc from 71.3 to 72.1, and MRMR from 48.0 to 49.8. This validates that a moderate fixed-capacity bottleneck with learnable reasoning tokens enables effective retrieval without explicit CoT. Further increasing $K$ to 16 yields only marginal  gains in MRMR while degrading standard retrieval performance, indicating excessive capacity weakens the bottleneck constraint. Notably, the throughput merely drops slightly from $6.5$ to $6.1$ samples/s, verifying the inference efficiency of single-forward latent reasoning.

\paragraph{Ablation on Different IB Head.}
As shown in Table~\ref{tab:abla_head}, Qwen3-0.6B decoder head reaches an overall score of 64.4, only 0.1 higher than Qwen2.5-0.5B, whereas Qwen3-1.7B drops to 64.1. This reveals decoder capacity does not monotonically benefit bottleneck supervision, with larger decoders incurring higher training costs and longer supervision chains. For embedding heads, mean pooling scores 64.4 and surpasses attention pooling at 63.8, proving simple aggregation more effectively compresses bottleneck information. The results verify the efficacy of our lightweight decoder and aggregated embedding head design.

\begin{table}[!t]
\caption{Ablation on different head architectures.}
\centering
\renewcommand{\arraystretch}{1.05}
\resizebox{\columnwidth}{!}{
\begin{tabular}{ll ccc c}
\toprule
\textbf{Dec. Head} & \textbf{Emb. Head}
& \textbf{Image} & \textbf{Video} & \textbf{VisDoc} & \textbf{All} \\
\midrule
\multicolumn{6}{l}{\textit{Decoder Head variants (Emb. Head fixed)}} \\
Qwen2.5-0.5B & Mean Pooling & 69.0 & 44.2 & 71.8 & 64.3 \\

\rowcolor[HTML]{FFF0EB}
Qwen3-0.6B   & Mean Pooling & \textbf{69.3} & \textbf{44.5} & \textbf{72.1} & \textbf{64.4} \\

Qwen3-1.7B   & Mean Pooling & 69.0 & 44.2 & 71.8 & 64.1 \\

\midrule
\multicolumn{6}{l}{\textit{Embedding Head variants (Dec. Head fixed)}} \\
Qwen3-0.6B & Attn Pooling  & 68.7 & 43.8 & 71.0 & 63.8 \\

\rowcolor[HTML]{FFF0EB}
Qwen3-0.6B & Mean Pooling  & \textbf{69.3} & \textbf{44.5} & \textbf{72.1} & \textbf{64.4} \\

\bottomrule
\end{tabular}
}
\label{tab:abla_head}
\end{table}

\subsection{Analysis on Latent Reasoning}

Figure~\ref{fig:demo} shows two examples of how latent reasoning tokens enable deep thinking in a single forward pass.
In Context~1, given a motorcycle image and the question, the decoder reconstructs ``You can use this for racing,'' showing that latent reasoning activates world knowledge.
In Context~2, a geometric query asks for the slant height of a pyramid. The decoder generates ``The slant height of the pyramid is 4.3.'' indicating internalized mathematical reasoning such as the Pythagorean theorem.
The similarity comparisons of reasoning latent embeddings also suggest the same conclusion.

\section{Conclusion}

We present LaME, which frames embedding-oriented latent reasoning as an information bottleneck problem.
LaME achieves single-pass latent reasoning via learnable reason tokens,
requiring no supervision on the structure or content of the latent thinking process. Only decode and contrastive losses guide the bottleneck.
Experiments on MMEB-v2 demonstrate state-of-the-art performance with significant inference speedups over CoT and iterative methods.
Future work will explore adaptive bottleneck capacity and extension to broader and more complex retrieval tasks.

\section*{Limitations}


Although latent reasoning may be more likely to produce retrieval-friendly representations,
the lack of explicit CoT traces reduces interpretability to some extent compared to CoT-based methods.
Additionally, the two-stage training pipeline, while effective for stabilizing bottleneck optimization,
introduces extra engineering complexity compared to single-stage approaches.
Simplifying the training procedure without sacrificing performance remains an open problem.


\section*{Ethics Statement}


This work studies multimodal retrieval and representation learning, and does not involve human subject recruitment, clinical data, or direct user interaction. As with other retrieval systems, our method may inherit biases, noise, and content imbalance from the source datasets. Although the rewrite-based design is intended to reduce redundant reasoning and semantic distortion, it can still produce imperfect or biased outputs; therefore, downstream use should include dataset curation and application-specific safety filtering.

\section*{Acknowledgments}

This work was supported in part by the National Natural Science Foundation of China under Grant No.~62472399.

\bibliography{custom}

\clearpage
\appendix
\section{More Training Details}

\subsection{Prompts for Latent Decoder}
\label{sec:appendix_decoder_prompt}

\begin{figure}[!t]
    \centering
    \includegraphics[width=\columnwidth]{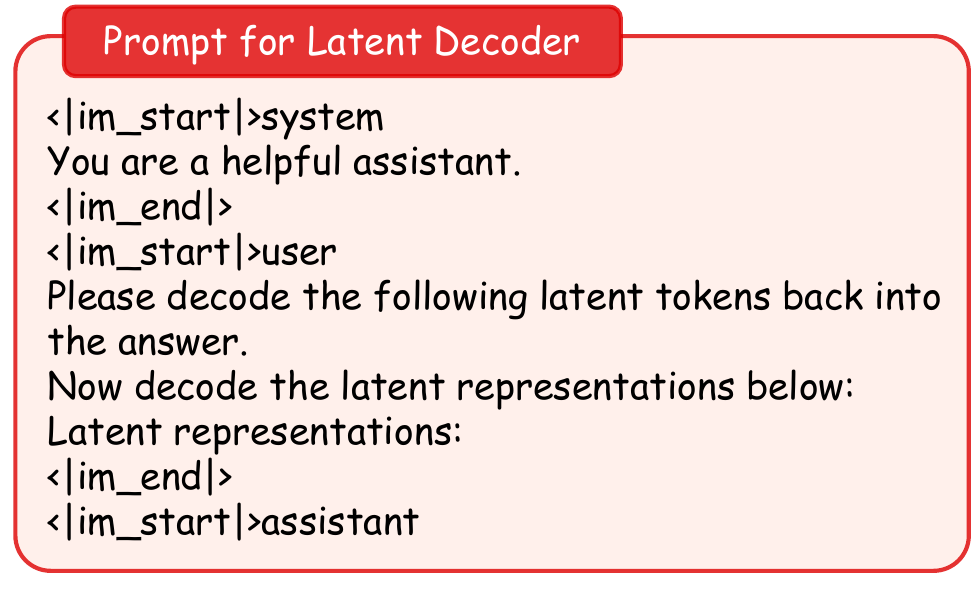}
    \caption{
    The prompt used by the latent decoder to reconstruct the target answer.
    }
    \label{fig:prompt}
\end{figure}

As shown in Figure~\ref{fig:prompt}, the decode prompt $s$ follows a standard Qwen chat template:
projected latent representations $\mathbf{h}_r^{1:K_r}$ are inserted after the reserved ``Latent representations:'' placeholder,
and a lightweight decoder $\psi$ then autoregressively decodes the target answer $y$.
This prompt is used only during training.

\subsection{Training Data Composition}
\label{sec:appendix_data_composition}

The training data comprises approximately \textbf{1.55 million} training pairs
drawn from the VLM2VEC-v2~\cite{meng2025vlm2vecv2} training corpus,
spanning three modalities: images, videos, and visual documents.
As shown in Table~\ref{tab:cot_filter_stats}, we report the detailed per-dataset statistics.
For image data, we adopt the image datasets in VLM2VEC-v2 including MSCOCO, ImageNet-1K, ChartQA, DocVQA, CIRR, and others across image-text matching, VQA, and retrieval scenarios ($\sim$677K samples after filtering).
For video data, we use LLaVA-Hound~\cite{zhang2024llavahound} with video QA, retrieval, and caption subsets ($\sim$744K samples).
For visual documents, we employ ViDoRe~\cite{faysse2024colpali} and VisRAG~\cite{yu2024visrag} covering complex charts, tables, and figures ($\sim$131K samples).
Notably, the decoder head reuses the answer fields from existing cold-start CoT datasets~\cite{jiang2026embedrl} as weak supervision targets,
completely avoiding dependency on CoT annotations.

\begin{table}[!t]
    \centering
    \setlength{\tabcolsep}{7pt}
    \small
    \caption{Statistics of training data composition.}
    \label{tab:cot_filter_stats}
    \resizebox{\linewidth}{!}{
        \begin{tabular}{@{}ccccc@{}}
            \toprule
            \textbf{Dataset} & \textbf{Initial} & \textbf{Filtered} & \textbf{Ratio} & \textbf{Modality} \\
            \midrule
            \rowcolor[HTML]{FFF0EB}
            \multicolumn{5}{c}{\textit{Image-based (MMEB-train)}} \\
            A-OKVQA & 50,000 & 34,750 & 69.50\% & Text-Image → Text \\
            CIRR & 50,000 & 31,950 & 63.90\% & Text-Image → Text-Image \\
            ChartQA & 50,000 & 35,900 & 71.80\% & Text-Image → Text \\
            DocVQA & 50,000 & 43,050 & 86.10\% & Text-Image → Text \\
            HatefulMemes & 25,500 & 15,150 & 59.41\% & Text-Image → Text \\
            ImageNet-1K & 50,000 & 40,200 & 80.40\% & Text-Image → Text \\
            InfographicsVQA & 50,000 & 36,850 & 73.70\% & Text-Image → Text \\
            MSCOCO & 50,000 & 23,800 & 47.60\% & Text-Image → Text-Image \\
            MSCOCO-i2t & 50,000 & 42,300 & 84.60\% & Text-Image → Text \\
            MSCOCO-t2i & 50,000 & 39,300 & 78.60\% & Text → Text-Image \\
            N24News & 50,000 & 27,700 & 55.40\% & Text-Image → Text \\
            NIGHTS & 47,823 & 39,300 & 82.17\% & Text-Image → Text-Image \\
            OK-VQA & 27,027 & 18,150 & 67.16\% & Text-Image → Text \\
            SUN397 & 50,000 & 41,700 & 83.40\% & Text-Image → Text \\
            VOC2007 & 23,532 & 18,600 & 79.05\% & Text-Image → Text \\
            Visual7W & 50,000 & 37,950 & 75.90\% & Text-Image → Text \\
            VisDial & 50,000 & 31,500 & 63.00\% & Text → Text-Image \\
            VisualNews-i2t & 50,000 & 31,300 & 62.60\% & Text-Image → Text \\
            VisualNews-t2i & 50,000 & 26,000 & 52.00\% & Text → Text-Image \\
            WebQA & 50,000 & 39,900 & 79.80\% & Text → Text-Image \\[2pt]
            
            \rowcolor[HTML]{FFF0EB}
            \multicolumn{5}{c}{\textit{Video-based (LLaVA-Hound)}} \\
            Caption Retrieval & 300,000 & 258,200 & 86.07\% & Video → Text \\
            Video QA & 300,000 & 249,200 & 83.07\% & Video-Text → Text \\
            Video Retrieval & 300,000 & 236,900 & 78.97\% & Text → Video \\[2pt]
            
            \rowcolor[HTML]{FFF0EB}
            \multicolumn{5}{c}{\textit{Document-based}} \\
            ViDoRe & 100,000 & 76,600 & 76.60\% & Text-Image → Text \\
            VisRAG & 100,000 & 54,850 & 54.85\% & Text → Image \\
            
            \midrule
            \textbf{Image-based} & 1,123,882 & 677,350 & 60.27\% & Image-centric \\
            \textbf{Video-based} & 900,000 & 744,300 & 82.70\% & Video-centric \\
            \textbf{Document-based} & 200,000 & 131,450 & 65.72\% & Document-centric \\
            \textbf{Total} & \textbf{2,223,882} & \textbf{1,553,100} & \textbf{69.84}\% & Multimodal \\
            \bottomrule
        \end{tabular}
    }
\end{table}

\section{More Experiments and Analysis}

\subsection{Ablation on Decoding Strategy}
\label{sec:appendix_ablation_decoding}

\begin{table}[!t]
\centering
\caption{Ablation on decoding strategies on MMEB-V2.}
\label{tab:abla_decode}
\setlength{\tabcolsep}{10pt}   
\renewcommand{\arraystretch}{1.1}
\resizebox{\columnwidth}{!}{
\begin{tabular}{l c c c c}
\toprule
\textbf{Decoding Strategy} & \textbf{Image} & \textbf{Video} & \textbf{VisDoc} & \textbf{All} \\
\midrule
No Decoder (Emb. Only) & 68.5 & 43.9 & 71.3 & 63.8 \\
\rowcolor[HTML]{FFF0EB}
Only-Answer            & \textbf{69.3} & \textbf{44.5} & \textbf{72.1} & \textbf{64.4} \\
Only-Thinking           & 68.4 & 43.5 & 71.4 & 63.7 \\
Chain-of-Thought                    & 68.4 & 44.1 & 71.4 & 63.8 \\
\bottomrule
\end{tabular}
}
\end{table}

As shown in Table~\ref{tab:abla_decode}, Only-Answer achieves the best results (69.3 / 44.5 / 72.1),
surpassing both CoT decoding and No Decoder (both 63.8 avg).
This highlights a critical issue: in long CoT sequences, abundant easy-to-predict tokens (e.g., function words)
rapidly saturate the LM loss and dominate gradients, thereby weakening IB supervision.
The bottleneck collapses into a generic average representation suited for trivial prediction but lacking discriminative power.
Only-Answer avoids this by restricting supervision to concise, informative answer tokens, preserving IB's role in encoding discriminative cues.
No Decoder's degradation further confirms the decoder provides essential training-time reasoning supervision despite being discarded at inference.

\subsection{Reason \textit{vs.} Final Retrieval Embedding}
\label{sec:appendix_reason_vs_embedding}

LaME produces two types of embeddings:
the reason embedding $\mathbf{e}$, derived from the reason tokens $\mathbf{h}_r^{K_r+1:K}$,
and the final retrieval embedding from the [EMBED] token.
We compare them across all modalities in Table~\ref{tab:embed_compare}.
The reason embedding achieves a competitive 64.1 overall score,
trailing the final retrieval embedding by merely 0.4 points.
This small gap indicates that $\mathcal{L}_{\text{Emb}}$ effectively trains the reason tokens to encode retrieval-relevant information. 
Notably, the gap widens to 0.9 on VisDoc, compared to 0.2 on Image and 0.4 on Video,
suggesting that fine-grained document understanding benefits more from the [EMBED] token's access to the complete hidden state sequence,
whereas compressed reason tokens may lose granular details required for dense visual-text alignment.

\subsection{Ablation on the Diversity Regularizer}
\label{sec:appendix_diversity}

Table~\ref{tab:abla_div} reports the results with varying diversity regularizer weight $\lambda_{\text{Div}}$.
Removing the regularizer ($\lambda_{\text{Div}} = 0$) yields the lowest score of 63.4.
Performance improves consistently as $\lambda_{\text{Div}}$ increases from 0 to 0.05, reaching the best overall result of 64.4.
These results confirm that the diversity regularizer is essential to prevent the reason tokens from \textit{collapsing} into degenerate representations during training,
thereby preserving the expressiveness of the latent reasoning process.
\section{Theoretical Analysis of Information Bottleneck}
\label{sec:appendix_info_bottleneck}

\begin{table}[!t]
\centering
\caption{Comparison of reason embedding and final retrieval embedding on MMEB-V2.}
\label{tab:embed_compare}
\setlength{\tabcolsep}{12pt} 
\renewcommand{\arraystretch}{1.1}
\resizebox{\columnwidth}{!}{
\begin{tabular}{l c c c c}
\toprule
\textbf{Embedding Source} & \textbf{Image} & \textbf{Video} & \textbf{VisDoc} & \textbf{All} \\
\midrule
Reason  & 69.1 & 44.1 & 71.2 & 64.0 \\
\rowcolor[HTML]{FFF0EB}
Final Retrieval & \textbf{69.3} & \textbf{44.5} & \textbf{72.1} & \textbf{64.4} \\

\bottomrule
\end{tabular}
}
\end{table}

\begin{table}[!t]
\centering
\caption{Effect of the diversity regularizer weight $\lambda_{\text{Div}}$ on MMEB-V2.}
\label{tab:abla_div}
\setlength{\tabcolsep}{18pt}  
\renewcommand{\arraystretch}{1.1}
\resizebox{\columnwidth}{!}{
\begin{tabular}{c c c c c}
\toprule
$\lambda_{\text{Div}}$ & \textbf{Image} & \textbf{Video} & \textbf{VisDoc} & \textbf{All} \\
\midrule
0.00  & 68.2 & 43.5 & 70.8 & 63.4 \\
0.02  & 69.0 & 44.1 & 71.6 & 64.1 \\
\rowcolor[HTML]{FFF0EB}
0.05  & \textbf{69.3} & \textbf{44.5} & 72.1 & \textbf{64.4} \\
0.10  & 69.0 & 43.9 & \textbf{72.5} & 64.2 \\
\bottomrule
\end{tabular}
}
\end{table}
We provide a more formal treatment of the information bottleneck formulation in LaME.
Let $X$ denote the input (interleaved text and visual tokens),
$Z$ the representation carried by the reason token hidden states $\mathbf{h}_r$,
and $Y$ the supervision signal from both decoder and embedding heads.
The IB objective is:
\begin{equation}
\min_{p(z|x)} \; I(Z; X) - \beta\, I(Z; Y),
\label{eq:ib_general}
\end{equation}
where $\beta > 0$ trades off compression against predictive power.

In LaME, the capacity constraint is structural rather than variational:
the representation $Z$ is supported on at most $K$ tokens of dimensionality $d$,
imposing a hard information ceiling
$I(Z; X) \leq K \cdot d \cdot \log(1 + \text{SNR})$,
where SNR denotes the effective signal-to-noise ratio of the hidden states.
This is distinct from variational IB~\citep{alemi2017deep},
which replaces $I(Z; X)$ with a KL-divergence upper bound requiring an auxiliary prior and Monte Carlo estimation;
here, the token count $K$ itself serves as the bottleneck radius.

The dual-head supervision provides the predictive term $I(Z; Y)$ via two complementary channels:
the decoder head maximizes $I(Z_r; Y_{\text{dec}})$ through autoregressive reconstruction on the first $K_r$ tokens,
while the embedding head maximizes $I(Z_e; Y_{\text{emb}})$ through contrastive learning on the remaining $K_e$ tokens.
Since both heads share the same bottleneck, the finite capacity enforces a joint compression trade-off between reconstructive and discriminative information.
This prevents degeneration into either a reconstructive autoencoder that ignores retrieval structure or a collapsed embedding that discards fine-grained semantics.

The two-stage training can be interpreted as progressively tightening the bottleneck:
Stage~1 (frozen backbone) allows the reason tokens to discover a compact representation without altering the backbone,
while Stage~2 (joint optimization) enables the backbone to co-adapt, routing task-relevant information through the reason tokens while preserving encoding capacity in the [EMBED] token path.
This staged curriculum avoids the trivial solution where the backbone bypasses the bottleneck entirely, which occurs under end-to-end training as shown in Table~\ref{tab:abla_com}.
%
%
%


\end{document}